\documentclass[11pt,a4paper]{article}
\usepackage{times,latexsym}
\usepackage{url}
\usepackage[T1]{fontenc}
\usepackage{amsmath}
\usepackage{amssymb}
\usepackage{framed}
\usepackage{graphicx}
\usepackage{xcolor}
\usepackage{bm}
\usepackage{wasysym}
\usepackage{enumitem}
\usepackage[utf8]{inputenc}
\usepackage[russian, english]{babel}
\usepackage[T2A, T1]{fontenc}
\usepackage[noline,noend, boxruled]{algorithm2e}
\usepackage{hyperref}

\usepackage[acceptedWithA]{tacl2018v2}

\usepackage{xspace,mfirstuc,tabulary}

\newcommand{\commentout}[1]{}

\newif\iftaclinstructions
\taclinstructionsfalse 
\iftaclinstructions
\renewcommand{\confidential}{}
\renewcommand{\anonsubtext}{(No author info supplied here, for consistency with
TACL-submission anonymization requirements)}
\newcommand{\instr}
\fi

\iftaclpubformat 

\else

\fi

\title{A Biologically Plausible Parser}

\author{
 Daniel Mitropolsky \\ 
 Department of Computer Science \\
 Columbia University \\
 New York, NY 10027 \And
 Michael J. Collins \\
 Google Research \\
 New York, NY 10011 \And
 Christos H. Papadimitriou \\
 Department of Computer Science \\
 Columbia University \\
 New York, NY 10027
}

\date{}

\begin{document}
\maketitle
\begin{abstract}
We describe a parser of English effectuated by biologically plausible neurons and synapses, and implemented through the Assembly Calculus, a recently proposed computational framework for cognitive function.   We demonstrate that this device is capable of correctly parsing reasonably nontrivial sentences
\footnote{Code available \url{https://www.github.com/dmitropolsky/assemblies}.}.  While our experiments entail rather simple sentences in English, our results suggest that the parser can be extended beyond what we have implemented, to several directions encompassing much of language. For example, we present a simple Russian version of the parser, and discuss how to handle recursion, embedding, and polysemy. 
\end{abstract}

\section{Introduction}
Language is a distinguishing human function involving the creation, articulation, comprehension, and maintenance of hierarchically structured information about the world. It is beyond doubt that language is achieved through the activity of neurons and synapses --- but how?  
There has been extensive previous work in cognitive experiments --- psycholinguistics, computational psycholinguistics, and brain imaging --- that has led to many insights into how the brain processes language (see section \ref{related} for an overview).
However, no concrete narrative emerges yet from these advances about the precise way in which the activity of individual neurons can result in language.  In particular, {\em we are not aware of an experiment in which a reasonably complex linguistic phenomenon is reproduced through simulated neurons and synapses.} This is the direction pursued here. 

Developing an overarching computational understanding of the way neurons in the human brain can make language is hindered by the state of neuroscience, which (a) predominantly studies sensory and motor brain functions of animals other than humans; and (b) with respect to computational modeling, focuses on the level of neurons and neuronal circuits, and lacks the kind of high-level computational model that seems to be needed for understanding how high-level cognitive functions can emerge from neuronal activity.

Very recently, a computational model of brain function, the Assembly Calculus (AC), has been proposed \cite{PNAS}.  The AC describes a dynamical system involving the following parts and properties, well attested in the neuroscience literature: a) brain areas with random connections between neurons; b) a simple linear model of neuronal inputs; c) inhibition within each area such that the top $k$ most activated neurons fire; d) a simple model of Hebbian plasticity, whereby the strength of synapses increase as neurons fire. 
An important object emerges from these properties: the {\em assembly}, a large set of highly interconnected excitatory neurons, all residing in the same brain area. 

Assemblies have been hypothesized by Hebb seven decades ago \cite{Hebb}, and were identified in the brain of experimental animals two decades ago \cite{Harris}. 
There is a growing consensus that assemblies play a central role in the way brains work \cite{Eichenbaum}, and were recently called ``the alphabet of the brain'' \cite{buzsakibook}. Assemblies can, through their near-simultaneous excitation, represent an object, episode, word or idea. It is shown in \citet{PNAS} that assemblies are an emergent property of the dynamical system under conditions (a) -- (d) above, both in theory and in simulations.   

In the AC (reviewed in more detail in Section 3), the dynamical system also makes it possible to create and manipulate assemblies through {\em operations} like projection, reciprocal projection, association, pattern completion, and merge. 
These operations are {\em realistic} in two orthogonal senses: first, they correspond to behaviors of assemblies that were either observed in experiments, or are helpful in explaining other experiments. Second, they provably correspond (they ``compile down'') to the activity of individual neurons and synapses; in \citet{PNAS}, this correspondence is proven both mathematically and through simulations. It is hypothesized in \citet{PNAS} that the AC may underlie high-level cognitive functions.  In particular, in the discussion section it is proposed that one particular operation of the AC called {\em merge} may play a role in the generation of sentences.

Note that, regarding the soundness of the AC, what has been established through mathematical proof is that AC operations work correctly {\em with high probability,} where the underlying probabilistic event is the creation of the random connectome.  So far, the simulation experiments conducted with the AC have demonstrated that individual commands of the AC, or short sequences thereof, can be successfully and reliably implemented in a simulator \cite{PNAS}.  Do such simulations scale?  For example, {\em can one implement in the AC a computationally demanding cognitive function, such as the parsing of sentences, and will the resulting dynamical system be stable and reliable?}  This is nature of the experiment we describe below.


{\em In this paper we present a Parser powered by the AC}.  In other words, we design and simulate a biologically realistic dynamical system involving stylized neurons, synapses, and brain areas. We start by encoding each word in the language as a different {\em assembly of neurons} in the brain area we call {\sc Lex}. Next, we feed this dynamical system with a sequence of words (signals that excite the corresponding sequence of word-coding assemblies in {\sc Lex}).  

The important questions are: {\em Will the dynamical system parse sentences correctly?  And how will we know?}

To answer the last question first, our dynamical system has a Readout procedure which, after the processing of the sequence, revisits all areas of the system and recovers a linked structure. We require that this structure be a parse of the input sentence.  Our experiments show that our Parser can indeed parse correctly, in the above sense, reasonably nontrivial sentences, and in fact do so at a speed  (i.e., number of cycles of neuron firings) commensurate with that of the language organ. 

Our design of this device engages several brain areas, as well as fibers of synapses connecting these areas, and uses the operations of the AC, enhanced in small ways explained in Section 3. It would in principle be possible to use the original set of AC operations in \citet{PNAS}, but this would complicate its operation and entail the introduction of more brain areas\footnote{``Brain areas'' is used here in the sense of the AC explained in Section 3, and does not necessarily correspond to significant accepted parts of the brain, such as the Brodmann Areas.}.  The resulting device relies on powerful word representations, and is essentially a lexicalized 
parser, producing something akin to a dependency graph.  

While our experiments entail the parsing of rather simple sentences in English, in Section \ref{extensions} we argue that our Parser can potentially be extended in many diverse directions such as error detection and recovery, polysemy and ambiguity, recursion, and languages beyond English. We also build a toy Russian parser, as well as a universal device that takes as its input a description of the language in the form of word representations, syntactic actions and connected brain areas. 

\paragraph{Goals.}  This research seeks to explore the two questions already highlighted above:
\begin{enumerate}
\item {Can a reasonably complex linguistic phenomenon, such as the parsing of sentences, be implemented through simulated neurons and synapses?}
\item {Can a computationally demanding cognitive function be implemented by the AC, and will the resulting dynamical system be stable and reliable?}
\end{enumerate}
In particular, we are not claiming that our implementation of the Parser necessarily resembles the way in which the brain actually implements language, or that it can predict experimental data. Rather, we see the parser as an existence proof of a nontrivial linguistic device built entirely out of simulated neuron dynamics.

\section{Related Work} \label{related}
\paragraph{Computational psycholinguistics.} There is a rich line of work in computational psycholinguistics on cognitively plausible models of human language processing. Such work focuses chiefly on (a) understanding whether high-level parsing methods can be used to predict psycholinguistic data (such as reading-time or eye-tracking data) and neural data (eg. fMRI and ECoG data from linguistic experiments); and (b) developing parsing methods that have specific, hallmark, experimentally-established cognitive properties of human syntactic processing (most importantly {\em incrementality} of parsing, {\em limited memory} constraints, and {\em connectedness} of the syntactic structures maintained by the parser). See \citet{keller-survey} for a summary of the psycholinguistic desiderata of (b), as well as a discussion of evaluation standards for (a). 
Exemplars of this line of work are Jurafsky’s use of probabilistic parsing to predict reading difficulty \cite{jurafsky}, the surprisal-based models of Hale, Levy, Demberg and Keller \cite{Hale2001, Levy2008, DembergKeller2008a}, and the strictly incremental predictive models of Demberg and Keller \cite{DembergKeller2008b, DembergKeller2009}.  Much work attempts to achieve the properties in (b) above while maintaining (a), that is, neural and psycholinguistic predictiveness, e.g.~the PLTAG parser of \citet{DembergPLTAG}, or the parser of  \citet{VasisthLewis}, which is constructed in ACT-R, a high-level meta-model of cognitive processing. 

A related line of work takes modern neural (ANN) parsing methods, and examines whether the internal states of these models at parse time can be used to model psycholinguistic and neurological data observed for the same sentences --- see \citet{findingsyntax} which uses an neural action-based parser, \citet{FrankRNNTransformer}  and \citet{FrankRNNReading} which examine various RNN architectures and transformers, and \citet{Schrimpf} which compares a wide range of ANN methods. A related direction is that of \citet{Hinaut} who build a {\em reservoir} network (a recurrent network that builds a representation of a sentence using fixed, random sparse matrices for weights) and build a classifier on top that predicts grammatical roles and that has also has some psycholinguistic predictivity (this has some similarities to aspects in Assembly Calculus, namely the sparse, random connections).

The present paper differs from these works in key ways.  In all previous work, parsers are written in a high-level programming language, whereas we focus on whether a simple parser can be implemented by millions of individual neurons and synapses through the simulation of a realistic mathematical model of neuronal processing.  In this framework, it is nontrivial to implement a single elementary step of syntactic processing, such as recording the fact that a particular input word is the sentence's subject, whereas in previous work such actions are built into the framework's primitives. In a survey paper of cognitive models of language that use ANNs, Frank summarizes that {\em "In spite of the superficial similarities between artificial and biological neural networks ... these cognitive models are not usually claimed to
simulate processing at the level of biological neurons. Rather ... neural network models form a description at Marr's algorithmic level"} \cite{FrankSurvey}. Our work can be seen as largely orthogonal to the related work described above, as we attempt to bridge granular neuronal mechanics with the study of complex cognitive processes such as syntax. See Section \ref{extensions} for discussions of potential future work that connects the two areas.

\paragraph{Neuroscience.} In building a neural parser, we begin with basic, established tenets of neuron biology: individual neurons fire when they receive sufficient excitatory input from other neurons; firing is an atomic operation; and some synapses can be inhibitory (firing one neuron decreases the total synaptic input of another neuron). We also assume a simplified narrative of synaptic plasticity: connections between neurons become stronger with repeated co-firing (Hebbian plasticity); this is a well-known abstraction of the many kinds of plasticity in the brain.  These tenets are covered in a number of textbooks (see for instance \citet{NeuralScience}, Chapters 7, 8, and 67). Our parser is built on a mathematical formalization of these principles.

How are higher-level cognitive processes computed by networks of individual neurons? Highly interconnected sets of neurons, called {\em assemblies}, are an increasingly popular hypothesis for the main unit of higher-level cognition in modern neuroscience. First hypothesized decades ago by Hebb, assemblies have been identified experimentally \cite{Harris} and their dynamics have been studied in animal brains (see e.g.~\citet{MillerETAL:14}), displacing previously dominant theories of information encoding in the brain, see e.g.~\citet{Eichenbaum}. 

In a recent paper \cite{PNAS}, a concrete mathematical formalization of assemblies is proposed.  They demonstrate that this simplified model is capable of simulating assembly dynamics that are known experimentally. This model of neurons, assemblies, and their dynamics can be viewed as a computational system, called the Assembly Calculus (AC), bridging neuronal dynamics with cognition; this is the computational system in which we implement our parser. The AC is summarized in Section 3. Note that it has been long debated whether language is a highly specialized system or is based in general cognitive faculties (see, e.g., the summary of the debate in \citet{VasisthLewis}). We are agnostic in this debate, because assemblies are the proposed unit of neural computation both specialized and generic, having been studied across a variety of systems and species \cite{MillerETAL:14,carrillo2018triggering}.

\paragraph{Language in the brain.}  
The AC model makes use of abstracted {\em brain areas} (defined in Section 3)
; therefore the design of our parser starts with the identification of a set of such areas.  Here we discuss how our choices relate to what is known about language in the brain --- a field in which, as of now, much is yet unknown or debated.

It has been known for 150 years that Wernicke's area in the superior temporal gyrus (STG) and Broca's area in the frontal lobe are involved in language; it is now generally --- but not universally, see the discussion below --- accepted that Broca's area is implicated in the processing of syntactic structures, while Wernicke's area is involved in word use. Language processing appears to start with access to a {\em lexicon}, a look-up table of word representations thought to reside in the left medial temporal lobe (MTL) opposite Wernicke's area. Major anatomical and cytological differences are known between humans and chimps at and near those areas of the left hemisphere, with evolutionary novel powerful fibers of synapses connecting these areas in humans \cite{Schomers}. Based on this, we believe the inclusion of a lexicon brain area ({\sc Lex}), containing assemblies representing words, is largely uncontroversial, as are its strong synaptic connections into other brain areas used by the parser. 

The book \citet{friederici} provides an excellent overview of a major direction in the theory of the language organ. After word look-up, activity in the STG is thought to signify the identification of syntactic roles; for example, it is known that the same noun is represented at different points in STG when it is the subject vs. the object of a sentence \cite{Frankland}, suggesting that there are specialized areas representing subject, object, presumably verb, and perhaps other syntactic categories. However, there is active discussion in the literature on whether brain areas are dedicated to specific linguistic functions such as syntactic and semantic processing, see for example \citet{Fedorenko2020,FedorenkoNetworks,Pylkkanen}. In our parser, we do make use of areas for different syntactic roles, but in doing so, we are not taking sides in the debate over the syntactic specialization of brain areas; we are not claiming that syntactic analysis is the exclusive function of these areas --- even the LEX area containing representations of words could be part of a larger memory area in the medial temporal lobe partaking in several aspects of language. 

At the highest level, 
the parser is generating a hierarchical dependency-based structure of a sentence that is processed incrementally. In the brain, creation of phrases or sentences seems to activate Broca's area --- what in \citet{ZaccarellaF} is called ``merge in the brain.''  Long sequences of rhythmic sentences each consisting of 4 monosyllabic words of the same syntactic structure as in ``bad cats eat fish,'' dictated at 4 Hz, result in brain signals from the subject's brain with Fourier peaks at 1, 2, and 4 Hz, suggesting that the brain is indeed forming hierarchical structures \cite{Poeppel}. Our parser represents a plausible hypothesis for the mechanism behind this process, in that it implements it within a realistic model of neural computation.  

\section{The Assembly Calculus} This section describes the version of the assembly calculus (AC) used in our simulator and experiments. This version of the AC is almost the same as that of the previous work of \citet{PNAS}, but includes minor modifications described here.

The AC is a computational system intended to model cognitive function in a stylized yet biologically plausible manner, by providing a high-level description and control of a {\em dynamical system} of firing neurons. 
There is a finite number $a$ of brain {\em areas} $A, B,...$ each containing $n$ excitatory neurons.  The $n$ neurons in each area are connected by a random weighted directed $G_{n,p}$ graph, meaning that every ordered pair of neurons has, independently, the same probability $p$ of being connected.  Each synapse $(i,j)$ has a synaptic weight $w_{ij}>0$, initialized to $1$, which changes dynamically.  For certain {\em unordered} pairs of areas $(A,B), A\neq B$, there is a random directed {\em bipartite} graph connecting neurons in $A$ to neurons in $B$ and back, again with probability $p$ for each possible synapse.  These connections between areas are called {\em fibers}.  All said, the dynamical system is a large dynamically weighted directed graph $G=(N,E)$ with $an$ nodes and random directed weighted edges.

Events happen in discrete time steps (think of each step as  20 ms).  The {\em state} of the dynamical system at each time step $t$ consists of (a) for each neuron $i$ a bit $f_i^t\in \{0,1\}$ denoting whether or not $i$ {\em fires} at time $t$, and (b) the synaptic weights $w_{ij}^t$ of all synapses in $E$.  Given this state at time $t$, the state at time $t+1$ is computed as follows:
\begin{enumerate}
    \item For each neuron $i$ compute its {\em synaptic input} $SI_{i}^t=\sum_{(j,i)\in E, f_j^t=1}w_{ji}^t$, that is, the sum total of all weights from pre-synaptic neurons that fired at time $t$.
    \item For each neuron $f_i^{t+1}=1$ --- that is, $i$ fires at time $t+1$ --- if $i$ is among the $k$ neurons {\em in its area} with the highest $SI_i^t$ (breaking any ties arbitrarily).
    \item For each synapse $(i,j)\in E$,\\
    $w_{ij}^{t+1}=w_{ij}^t(1+ f^t_i f^{t+1}_j \beta)$; that is, a synaptic weight increases by a factor of $1+\beta$ if and only if the post-synaptic neuron fires at time $t+1$ and the pre-synaptic neuron had fired at time $t$.
\end{enumerate}
We call the set of $k$ neurons in an area firing at time $t$ the {\em cap} of that area.

These are the equations of the dynamical system.  The AC also provides commands for high-level {\em control} of this system. An area can be {\em inhibited} (that is, the neurons in it prevented from firing, respectively), and {\em disinhibited} (cancel the inhibition, if it is currently in effect).  We also assume that fibers can be inhibited (be prevented from carrying synaptic input to other areas).  In the brain, inhibition is accomplished through {\em populations of inhibitory neurons}, whose firing prevents other neurons from firing. In fact, there may be multiple populations that inhibit an area or a fiber. We denote this command by inhibit$(A,i)$, which inhibits $A$ through the excitation of a population named $i$. Similarly, disinhibition {\em inhibits} a population of inhibitory neurons (such as $i$ above), which currently inhibit $A$, an operation denoted disinhibit$(A,i)$. If, for instance, we inhibit$(A,i)$ and also inhibit$(A,j)$, and then disinhibit$(A,j)$, $A$ is still inhibited (because of $i$).  Finally, we assume that a {\em fiber} $(A,B)$ can be similarly inhibited or disinhibited, denoted inhibit$((A,B),i)$ and disinhibit$((A,B),i)$.


We now define the {\em state} of this dynamical system at time $t$. The state contains the firing state of each neuron, edge weights $w_{ij}$, and inhibition information. If an area $A$ is inhibited at time $t$, and disinhibited at time $t'$, we assume that for all $i \in A$, $j \in (t+1) \ldots t', f^j_i = f^t_i$. That is, the firing state is maintained during the entire period of inhibition. 

A critical emergent property of the system is that of {\em assemblies}. An {\em assembly} is a special set of $k$ neurons, all in the same area, that are {\em densely interconnected} --- that is, these $k$ neurons have far more synapses between them than random, and these synapses have very high weights --- and are known to represent in the brain objects, words, ideas, etc.  


Suppose that at time $0$, when nothing else fires, we execute fire$(x)$ for a fixed subset of $k$ neurons $x$ in area $A$ (often these $k$ neurons will correspond to an assembly), and suppose that there is an adjacent area $B$ (connected to $A$ through a fiber) where no neurons currently fire.  Since assembly $x$ in area $A$ fires at times $0,1,2,\ldots$ (and ignoring all other areas), it will effect at times $1,2,\ldots$ the firing of an evolving set of $k$ neurons in $B$ --- a sequence of caps ---, call these sets $y^{1},y^{2},\ldots$.  At time $1$, $y^{1}$ will be simply the $k$ neurons in $B$ that happened to receive the largest synaptic input from $x$.  At time $2$, $y^{2}$ will be the set of neurons in $B$ that receive the highest synaptic input from $x$ {\em and} $y^1$ combined --- and recall that the synaptic weights from $x$ to $y^1$ have increased. If this continues, it is shown in \citet{PNAS} that, with high probability (where the probability space is the random connectivity of the system), the sequence $\{y^t\}$  eventually converges to a stable assembly $y$ in $B$, called {\em the projection of $x$ in $B$}. There are more high-level operations in $AC$ (reciprocal projection, merge, association, etc.), comprising a computational framework capable of carrying out arbitrary space-bounded computations. Here we shall only focus on projection, albeit {\em enhanced} as described below. 

Suppose an assembly $x$ in area $A$ is projected as above to area $B$ to form a new assembly $y$, a process during which neurons in $B$ fire back into $A$. It is quite intuitive that, after projection, $y$ is densely interconnected and it has dense synaptic connections from the neurons in $x$, because of the way in which it was constructed. Consequently, if $x$ ever fires again, $y$ will follow suit.  In this paper we also assume (and our experiments validate this assumption) that, since the fiber between $A$ and $B$ is reciprocal, there are also strong synaptic connections from $y$ to $x$, so that, if $y$ fires, $x$ will also fire next.


We next define an enhancement of the projection operation --- tantamount to a sequence of projection operations --- which we call {\em strong projection} (see Algorithm 1). Consider all disinhibited areas of the dynamical system, and all disinhibited fibers containing them.  This defines an undirected graph (which in our usage will always be a tree).  Call a disinhibited area {\em active} if it contains an assembly --- the one most recently activated in it.  Now, suppose that all these assemblies fire simultaneously, into every other disinhibited adjacent area through every disinhibited fiber, and these areas fire in turn, possibly creating new assemblies and firing further down the tree, until the process stabilizes (that is, the same neurons keep firing from one step to the next). We denote this system-wide operation {\em strong project} or project$^*$. Note that project$^*$ is {\em almost} syntactic sugar, as it simply abbreviates a sequence of projections (which can be done in the AC model); however, the notion of an {\em active} area that we use is a small addition to the AC. Though this modification is minor, it simplifies the parser implementation, but it could be removed at the expense of more AC brain areas and perhaps time steps. 

Our experiments with the Parser show that indeed the operation project$^*$ works as described --- that is, the process always stabilizes.   
We introduce the term {\em strong Assembly Calculus} (sAC) to refer to the computational system whose operations are: {\em inhibit} and {\em disinhibit}, applied to any area or fiber, and the strong projection operation {\em project$^*$}.  It is not hard to see sAC is Turing-complete in the same sense as AC is shown in \citet{PNAS}, but we shall not need this.

\begin{algorithm}
\BlankLine
\ForEach{\upshape area $A$}{
    \If{\upshape there is active assembly $x$ in $A$}{
    \ForEach{$i \in x$}{
        $f_i^1 = 1$ \;
    }
    }
}
\BlankLine
\For{$i=1,\ldots,20$}{
all $A$, all $i\in A$, initialize $SI_i^t = 0$ \;
\ForEach{\upshape uninhibited areas $A$, $B$}{
    \If{\upshape fiber $(A,B)$ inhibited}{
		skip \;
	}
	$x = \{i \in A : f_i^t = 1\}$ \;
	\ForEach{\upshape $j \in B$}{
		$SI_j^t$ += $\sum_{i \in x, (i,j)\in E} w_{ij}$ \;
	}
}
\BlankLine
\ForEach{\upshape uninhibited area $A$}{
	\ForEach{\upshape $i \in A$}{
	\eIf{\upshape $SI_i^t$ in top$-k_{j\in A}(SI_j^t)$ }{
			$f_i^{t+1}$ = 1 \;
	}{
		$f_i^{t+1}$ = 0 \;
	}
	}
}
\ForEach{\upshape uninhibited areas $A$,$B$}{
    \If{\upshape fiber $(A,B)$ inhibited}{
		skip \;
	}
	\ForEach{\upshape $(i,j) \in (A\times B)\cap E$}{
    \If{\upshape $f_i^t = 1$ and $f_j^{t+1}=1$}{
        $w_{ij}$ = $w_{ij} \times (1+\beta)$
    }
	}
}
}
\caption{AC code for project*}
\end{algorithm}

In the pseudocode above, the ``active assembly $x$ in $A$" means either the set of $k$ neurons most recently fired in $A$ (which, by the dynamics of the system, happens to be an assembly), or a set of $k$ neurons that is activated (set to fire at $t=1$) externally: we use this for activating the assembly corresponding to a fixed word in Section \ref{parser}.

\section{The Parser} \label{parser}

\subsection{Parser Architecture}
\begin{figure*} 
  \includegraphics[scale=0.9, trim=50 70 0 0, clip]{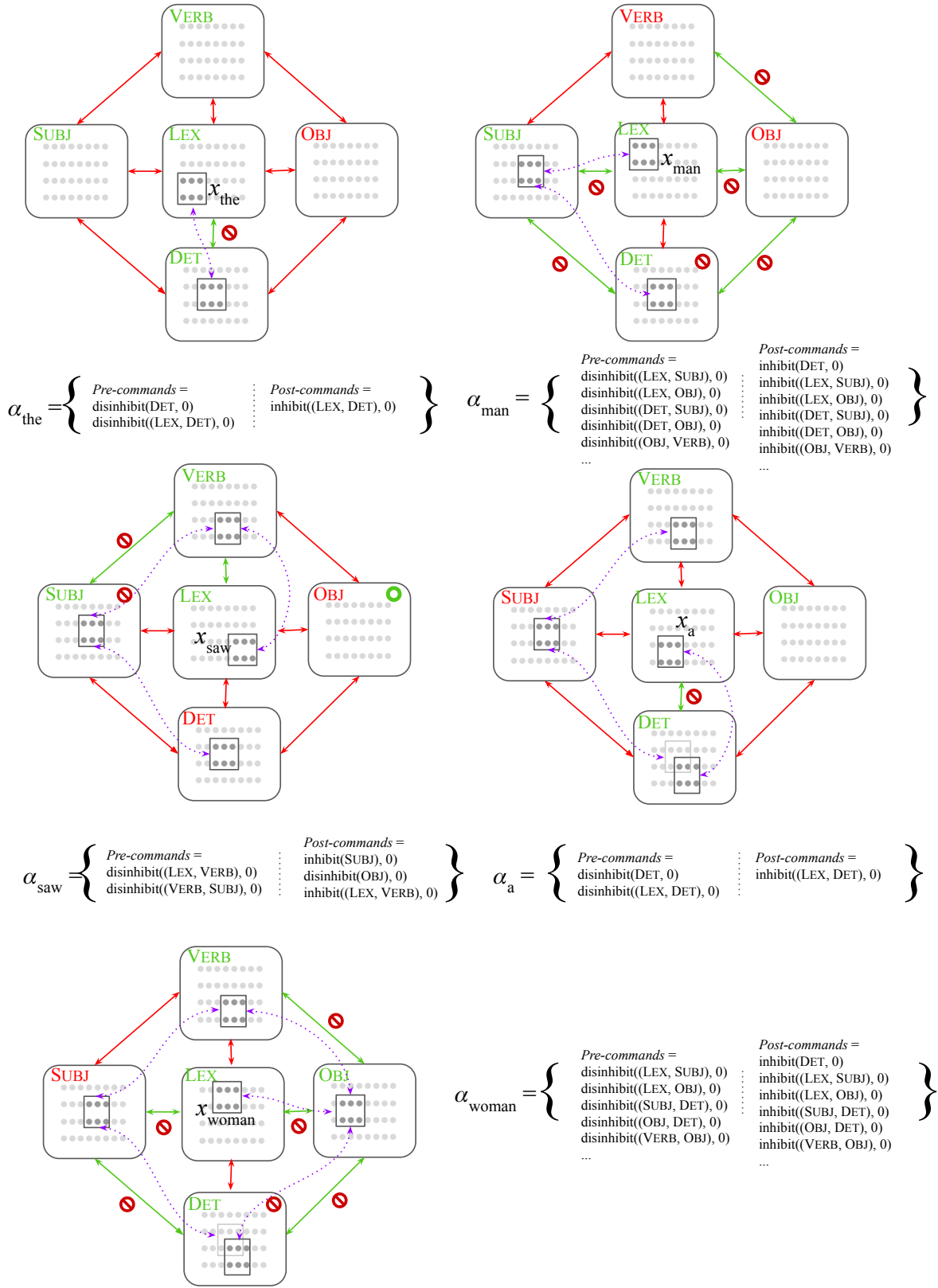}
  \caption{Trace of the Parser for the sentence {\em ``the man saw a woman"}. The action $\alpha_i$ of each word is given beneath a figure showing the state of the Parser after the Pre-commands and project* have occurred for that word. Green areas and fiber arrows are disinhibited; red are inhibited. All assemblies active right after project* are shown (with the one in {\sc Lex} labeled $x_w$). Purple dotted arrows indicate assemblies that have been connected; one can fire to activate the other. Purple arrows to {\sc Lex} are shown only in the stage in which they are created. Green ($\color{green}{\bm{\ocircle}}$) and red ($\color{red}\bm{\oslash}$) circles indicate areas or fibers that will be disinhibited/inhibited by the Post-commands.}
  \label{fig:big}
\end{figure*}

The Parser is a program in sAC whose data structure is an undirected graph $G=(\mathcal{A},\mathcal{F})$. $\mathcal{A}$ is a set of brain areas and $\mathcal{F}$ is a set of fibers, that is, unordered pairs of areas. One important area is {\sc Lex}, for {\em lexicon}, containing word representations.  {\sc Lex}, which in the brain is believed to reside in the left MTL, is connected through fibers with all other areas.  The remaining areas of $\mathcal{A}$ perhaps correspond to subareas of Wernicke's area in the left STG, to which words are projected from left MTL for syntactic role assignment.  In our experiments and narrative below these areas include {\sc Verb}, {\sc Subj}, {\sc Obj}, {\sc Det}, {\sc Adj}, {\sc Adv}, {\sc Prep}, and {\sc PrepP}. Besides {\sc Lex}, several of these areas are also connected with each other via fibers (see Figure \ref{fig:big}). Each of these areas was postulated because it seemed necessary for the Parser to work correctly on certain kinds of simple sentences.  As it turns out, they correspond, roughly yet unmistakably, to {\em dependency labels}. 
$\mathcal{A}$ can be extended with more areas, such as {\sc Conj} and {\sc Clause}; see Section \ref{extensions} for a discussion of some of these extensions. 

With the exception of {\sc Lex}, all of these areas are standard brain areas of the AC, containing $n$ randomly connected neurons of which at most $k$ fire at any point in time. In contrast, {\sc Lex} contains a fixed assembly $x_{{w}}$ for every word in the language. The $x_{{w}}$ assemblies are special, in that the firing of $x_{{w}}$ entails, besides the ordinary effects of firing assemblies have on the system, the execution of a short program $\alpha_w$ specific to word $w$, called the {\em action} of $w$.  Intuitively, the sum total of all actions in assemblies of {\sc Lex} constitute something akin to the device's {\em grammar.}

The action $\alpha_w$ of a word $w$ is two sets\footnote{We do not call them sequences, because we think of the commands in each as executed in parallel.} of {\sc Inhibit} and {\sc Disinhibit}
commands, for specific areas or fibers.  The first set is executed before a {\em project}* operation of the system, and the second afterwards.\footnote{To simplify notation, we {\em enumerate} the inhibitory populations $i$ separately for each area and fiber, i.e. each area and fiber can be inhibited by inhibitory populations $\{0,1,\ldots\}$; in most cases there is only one, rarely two.} 

\begin{figure}
    \centering
    \includegraphics[scale=0.8]{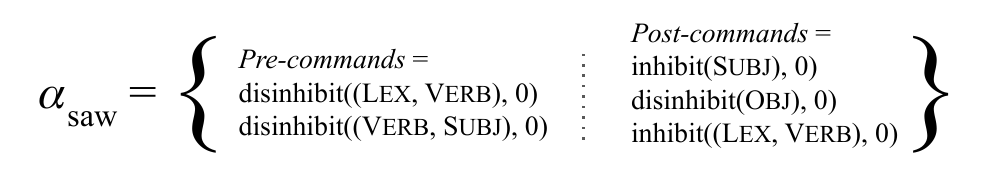}
    \caption{Action for transitive verb {\em saw}}
    \label{fig:action}
\end{figure}

The action for the word {\text chase} is shown in Figure \ref{fig:action} (More examples of actions and their commands for other parts of speech are given in figure \ref{fig:big}). In fact, every standard transitive verb has the same action. The pre-commands disinhibit the fibers from {\sc Lex} and {\sc Subj} to {\sc Verb}, allowing an assembly to be formed in {\sc Verb} that is the merge of the word assembly $x_{\text{chase}}$ in {\sc Lex} and the assembly representing the subject noun phrase in {\sc Subj}, which because subjects precede verbs in (our subset of) English, must have hitherto been constructed. Since {\em chase} is a transitive verb, the post-rules include the disinhibition of {\sc Obj} in anticipation of an obligatory object. In terms of neurons and synapses, we think that the $k$ firing neurons comprising the assembly $x_{{w}}$ contain certain neuron subpopulations whose firing excites the appropriate inhibitory and disinhibitory neural populations (the ones in the post commands through a delay operator).  These subpopulations may be shared between all transitive verbs.

\subsection{Operation} 
As shown in Algorithm 2, the Parser processes each word in a sentence sequentially. For each word, we activate its assembly in {\sc Lex}, applying the pre-commands of its lexical item. Then, we {\em project}*, projecting between disinhibited areas along disinhibited fibers. Afterwards, any post-commands are applied. 
\begin{algorithm}
\SetKwInOut{Input}{input}
\SetKwInOut{Output}{output}
\Input{a sentence $s$}
\Output{representation of dependency parse of $s$, rooted in {\sc Verb}}
\BlankLine
 disinhibit({\sc Lex}, 0) \;
 disinhibit({\sc Subj}, 0) \;
 disinhibit({\sc Verb}, 0) \;
 \ForEach{\upshape word $w$ in $s$}{
  activate assembly $x_w$ in {\sc Lex} \;
  \ForEach{\upshape pre-rule (Dis)inhibit($\square,i$) in $\alpha_w \rightarrow$ Pre-Commands}{
    (Dis)inhibit($\square, i$) \;
  }
  }
 {\em project}* \;  
 \ForEach{\upshape post-rule (Dis)inhibit($\square,i$) in $\alpha_w \rightarrow$ Post-Commands}{
    (Dis)inhibit($\square, i$)
  }
\BlankLine
 \caption{Parser, main loop}
\end{algorithm}

In the pseudocode above, $\square \in \mathcal{A}\cup \mathcal{F}$ can represent either an area or a fiber, depending on the command. 

\subsection{Readout}
After the completion of Algorithm 2, we will argue that a dependency tree of the sentence is represented implicitly in the synaptic connectivities $w_{ij}$ between and within brain areas of $G$. To verify this, we have a {\em readout} algorithm which, given the state of $G$ (in fact, only the $w_{ij}$ and the $k$ neurons last fired in {\sc Verb} are needed), outputs a list of dependencies. Our experiments verify that when applied to the state of $G$ after parsing a sentence with Algorithm 1, we get back the full set of dependencies. 

For notational convenience, we define an operation {\em try-project$(x,B)$}, in which an assembly $x$ in some area $A$ projects into area $B$, but only if this results in a stable assembly in $B$. This is tantamount to projecting only if $x$ was projected into $B$ during the Parser's execution (as part of a {\em project}* in some step). Lastly, define {\em getWord}() to be the function which, at any given time when an assembly $x_w$ is active in {\sc Lex}, returns the corresponding word $w$. In the following pseudocode, {\em try-project}$(x,B)$ returns an assembly $y$ in $B$ if it succeeds, and {\sc None} otherwise.

\begin{algorithm}
\SetKwInOut{Input}{input}
\SetKwInOut{Output}{output}
\Input{$G$ after parsing, with active root assembly $v$ in {\sc Verb}}
\Output{the parse tree stored implicitly in $G$}
\BlankLine
initialize stack $s$ as \{$(v,{\textsc Verb})$\} \;
initialize dependencies $\mathcal{D} = \{\}$ \;
 \While{\upshape $s$ not empty }{
  ($x$,$A$) = $s$.pop() \;
  {\em project}($x$, {\sc Lex}) \;
  $w_A$ = {\em getWord}() \;
  \ForEach{\upshape area $B \neq A$ s.t. $(A,B) \in \mathcal{F}$}{
  $y$ = {\em try-project}$(x,B)$ \;
  \If{$y$ not None}{
  {\em project}($y$, {\sc Lex}) \;
  $w_B$ = {\em getWord}() \;
  add dependency $w_A \xrightarrow{B} w_B$ to $\mathcal{D}$ \;
  s.insert($(y,B)$) \;
  }
  }
 }
 return $\mathcal{D}$ \;
 \caption{Read out of parse tree in $G$}
\end{algorithm}

We present the readout algorithm mostly as a means of proving the Parser works, since it demonstrates a mapping between the state of the Parser graph $G$ (namely, its synaptic weights) and a list of dependencies. However, we remark that it is not biologically implausible. {\em try-project} is not a new primitive, and can be implemented in terms of neurons, synapses, and firings. Most simply, $x$ can fire into $B$, and if the resulting $k$-cap in $B$ is not stable (changes with a repeated firing, say), it is not an assembly. Alternatively (and in simulation), one can project from $B$ into {\sc Lex}, and if the cap in $B$ is not a stable assembly formed during parsing, the $k$-cap in {\sc Lex} will {\em also} not be an assembly, i.e. not correspond to any valid $x_w$ (this is related to ``nonsense assemblies" discussed in Section \ref{error-detection}). Further, the stack of Algorithm 3 is only used for expository clarity; as the assemblies projected are always the ones projected to in the previous round, with some thought one could implement the algorithm with {\em project}*.

\section{Experiments}
We provide an implementation of the Parser in Python \footnote{The simulation code is available online; link not given to preserve anonymity.}. To achieve this, we have significantly extended the existing AC simulation library of \cite{PNAS} (for instance adding inhibition mechanics, and making possible brain areas like {\sc Lex} with fixed, pre-defined assemblies). Importantly, the entire algorithm, at its lowest level, is running by simulating {\em individual neurons and synapses}, and any higher-level concept or structure (such as the parse tree itself) is an abstraction that is encoded entirely in the individual firings and neuron-to-neuron synaptic weights of a giant graph and dynamical system representing the brain. 

We provide a set of 200 varied sentences that are parsed correctly by the Parser. For each sentence we have a target correct dependency structure; we verify that the Readout operation produces this dependency structure for each sentence. 

The sentences were designed by hand in order to demonstrate a variety of syntactic phenomena that our Parser (and the AC) can handle; this includes prepositional phrases, transitivity and intransitivity (including verbs that can be both transitive or intransitive), optional adverb placement pre- or post-verb, and several more. See Section \ref{appendix} for these example sentences and the syntactic pattern they represent. We remind the reader that to our knowledge, this is the first realistic parser implemented on the level of individual neurons and synapses.

The 200 sentences parsed were sampled from 10 syntactic patterns the Parser was designed to handle, from a vocabulary of 100 words.  Since the underlying dynamical system can in principle develop an instability with low probability, failure is still possible, but it did not happen in our experiments. There are syntactic structures that the Parser does not currently handle, but which we believe are possible in our model in two senses: first, the grammar (word actions) can be expanded to handle these structures, and more importantly, the dynamical system will still correctly parse them with high probability. A prime example of such a structure is sentence embedding (``the man said that..."). See Section \ref{extensions} for more discussion of such extensions. 

The speed of parsing in our simulations is commensurate with known bounds on neural language processing. In each step, the participating assemblies must fire enough times for the project$^*$ operation to stabilize; we find through simulation that for reasonable settings of the model’s hyperparameters, convergence occurs within 10-20 firing epochs, even when multiple brain areas are involved. Assuming 50-60 neuron spikes (AC time steps) per second, and allowing for the (parallel) execution of the inhibit/disinhibit operations in the word's action, which will take a small number of additional cycles, we arrive at a frequency range of 0.2-0.5 seconds/word.


\section{Extensions and Discussion} \label{extensions}
\paragraph{The UParser and a parser for Russian.}
The underlying meta-algorithm of the Parser, which we call the UParser, takes as input a set of areas $\mathcal{A}$, a set of fibers $\mathcal{F}$, a set of words $W$ and a set of actions $a_w$ for each $w \in W$ (whose commands manipulate only areas in $\mathcal{A}$), and parses with sAC using Algorithms 2 of Section 3. The UParser is language-agnostic, and can be seen as modeling a universal neural basis for language: a specific language instantiates this basis by specifying $\mathcal{A}$ (roughly, its syntactic constituents), $\mathcal{F}$, $W$ and the $\alpha_i$. This is in no way constrained to English; for instance, our model and algorithm are equally well equipped to handle a highly inflectional language with relatively free word order, where syntactic function is indicated morphologically. We demonstrate this with a parser for a toy subset of Russian, a language with free word order in  simple sentences. Particularly, our parser works on a set of sentences that is closed under permutation; permutations of the same sentence are parsed correctly, and produce the same dependency tree (as verified with Algorithm 3). See section \ref{appendix} for more details on the Russian Parser. 

\paragraph{The big bad problem and recursion.}
The Parser of Section 3 is rather skeletal and needs to be extended to handle recursive constructions such as compounding and embedded clauses. Consider the sentence “The big bad problem is scary”. Following the Parser of Section 3, after the word “big”, an assembly $a$ representing “big” has been created in {\sc Adj} which the system anticipates projecting into {\sc Subj} in a subsequent step (when a noun is encountered) to form an assembly in {\sc Subj} representing the subject NP. However, the next word is not a noun but “bad”, part of a chain of adjectives. Now, if we project from {\sc Lex} into {\sc Adj} to form an assembly $b$ representing “bad”, we lose all reference to $a$, and there is no way to recover ``big"! There are several simple and plausible solutions to the chaining problem, which can also be used to parse adverb chains, compound nouns, and other compound structures. 

One solution is using two areas {\sc Comp}$_1$ and {\sc Comp}$_2$ in $\mathcal{A}$. When we encounter the second adjective (or more generally a second word of the same part of speech which otherwise would fire into the same area, overwriting earlier assemblies), we project from {\sc Lex} into {\sc Comp}$_1$ instead of {\sc Adj}, but simultaneously project from {\sc Adj} into {\sc Comp}$_1$ to link the first adjective with the second. For a third adjective in the chain, we  project {\sc Lex} into {\sc Comp}$_2$ but also {\sc Comp}$_1$ into {\sc Comp}$_2$ {\em unidirectionally}; generally, for adjective $i$ in addition to {\sc Lex} we project unidirectionally from {\sc Comp}$_{\text{parity}(i)}$ (which contains the previous adjective assembly) into {\sc Comp}$_{\text{parity}(i-1)}$. We demonstrate parsing chains of varying lengths with these two areas in our simulations. However, one limitation of this solution is that it requires {\em unidirectional} fibers; if the projection from {\sc Comp}$_1$ to {\sc Comp}$_2$ above is reciprocal, it won't be possible to link the assembly in {\sc Comp}$_2$ to {\em another} assembly in {\sc Comp}$_1$.

Another approach, which may be more realistic and obviates the need for unidirectional fibers (though it cannot handle arbitrary-length chains) is to add more than 2 areas, {\sc Comp}$_1$,...,{\sc Comp}$_m$, for some small but reasonable $m$, perhaps $7$. Parsing proceeds as in the two area solution but by chaining {\sc Comp}$_i$ to {\sc Comp}$_{i+1}$. The number of areas $m$ could model well-studied cognitive limits in processing such complex structures, see \citet{GibsonEdward, karlsson}. To model longer chains, in high-demand situations or with practice, the brain could recruit additional brain areas.

Such maneuvers can handle right recursion of the form $S\rightarrow A^{*}!A^{+}$.  Center embedded recursion is more complicated. To construct a parse tree for a sentence with an embedded sentence or relative clause, for an execution of the Parser's inner loop, the subjects and/or the verbs of the two sentences may need to coexist in the corresponding areas.  This is not a problem, as brain areas can handle tens of thousands of assemblies, but the linking structure must be preserved.  It must be possible to save the state of the embedding sentence while the embedded one is serviced.  One possibility is to introduce {\sc CurrentVerb} areas, which can be chained like the compounding areas above, and will act as a stack of activation records for this more demanding form of recursion.  The idea is that, by recovering the verb of the embedding sentence, we can also recover, through linking and after the end of the embedded sentence, the full state of the embedding sentence at the time the embedded sentence had begun, and continue parsing.  This needs to be implemented and experimented with. 

\paragraph{Disambiguation.}
In English, {\em rock} can be an intransitive verb, a transitive verb, or a noun; so far, we have ignored the difficult problems of polysemy and ambiguity. To handle polysemy, every word $w$ may need to have many action sets $\alpha_w^1, \alpha_w^2,...$, and the parser must disambiguate between these.

We believe that the Parser can be extended to handle such ambiguity. The choice of the action could be computed by a {\em classifier}, taking as input a few words' worth of look-ahead and look-behind in the current sentence (or perhaps just their parts of speech), and selecting one of the action sets; the classifier can be trained on the corpus of previously seen sentences.  This also needs implementation and experimentation.

\paragraph{Error detection.} \label{error-detection}
Grammaticality judgment is the intuitive and intrinsic ability of native speakers of any language to judge well-formedness; this includes the ability to detect syntactic errors. Neurolinguists are increasingly interested in the neural basis of this phenomenon; for instance, recent experiments have detected consistent signals when a sentence fragment is suddenly continued in a way that is illegal syntactically \cite{chomsky-illusion}. Built into the Parser is the ability to detect some malformed sentences. 

There are at least two simple mechanisms for this. One is when a fragment is continued by a word that immediately makes it ungrammatical, such as following “the dogs lived” with “cats”. The Parser, having processed the intransitive verb “lived”, has not disinhibited area {\sc Obj}, and all other noun-related areas are inhibited. Upon encountering “cats”, {\em project}* will not fire any assemblies; we call this an  {\em empty-project} error. 

Other kinds of syntactic violations can be detected by {\em nonsense assembly} errors. Such an error occurs during readout when an area is projected into {\sc Lex}, but the resulting assembly in {\sc Lex} does not correspond to $x_w$ for any $w \in W$; in other words, when the function {\em getWord}() of the readout algorithm of Section 3 fails, which indicates that the state of $G$ must have resulted from an impossible sentence. We provide a list of illegal sentences for which our Parser simulation detects empty-project or nonsense-assembly errors, indicating different kinds of syntactic violations.

One kind of syntactic error our Parser does not detect currently is {\em number agreement:}  Our simulator would not complain on input ``Cats chases dogs;'' this is not hard to fix by having separate areas for the projection of singular and plural forms.  Other types of agreement, such as in gender, can be treated similarly.    

\paragraph{The role of Broca's area.}  Language processing, especially the formation of phrases and sentences, has been observed to activate Broca's area in the left frontal lobe; so far, the parser only models processes believed to be in the left STG and MTL. We hypothesize that Broca's area may be involved in the building of a concise syntax tree summarizing the sentence, consisting of only the subject, the verb, and the object if there is one (but with access to the rest of the parsed sentence through the other areas).  This would involve new areas {\sc S} (for sentence) and {\sc VP} (for verb phrase), with fibers from {\sc Verb} and {\sc Obj} to {\sc VP}, and from {\sc VP} and {\sc Subj} to {\sc S}.  Building the basic three-leaf syntax tree can be carried out passively and in the background while the rest of syntax processing is happening in our current model.

\paragraph{Closed classes and an alternative architecture.}  Words such as {\em the, of, my} may reside in areas dedicated to closed parts of speech in Wernicke's area, instead of being a part of {\sc Lex} as we assume, for simplicity, in our model and simulations. In fact, before exploring the architecture described here, we considered an alternative in this direction, which has been suggested in the neuroscience literature 
\cite{Rolls2015}.  

Suppose that {\sc Lex} in the left MTL only contains the phonological and morphological information necessary for recognizing words, and all grammatical information, such as the actions $\alpha_w$, reside in Wernicke's area, perhaps again in subareas not unlike the ones we postulate here.  For example, {\sc Subj} could contain every noun (in Russian, in its nominative form), as assemblies which are in permanent projection back and forth with the corresponding assemblies in {\sc Lex}; likewise for {\sc Obj} (accusative in Russian).  Verbs are all permanently projected in {\sc Verb}, and so on.  In this model, syntactic actions are specific to areas, not words.  We plan to further explore what new problems such a parser would present --- and which problems it would solve ---, as well as what advantages and disadvantages, in performance, complexity, and biological plausibility, it may have. 

\paragraph{Predictions of our Model.}  \label{predictions}
The Parser is essentially a proof of concept, an ``existence theorem'' establishing that complex cognitive, and in particular linguistic, functions can be implemented in a biologically realistic way through the Assembly Calculus.  To this end, we chose an architecture that is compatible with what is known and believed about language processing in large parts of the literature.  Several predictions can be made that can be verified through EEG and fMRI experiments.  One such prediction is that the processing of sentences of similar syntactic structure (for example, ``dogs chase cats'' and ``sheep like dogs'') would result in similar brain activity signatures, while sentences with a different syntactic structure (``give them food'') would generate different signatures.  We are currently working with CUNY cognitive neuroscientists Tony Ro and Tatiana Emmanouil to design and conduct such experiments. 

If and when extensive recording from large numbers of individual neurons in humans becomes possible, the Parser model can be concretely tested, as it predicts long-term assemblies corresponding to lexical items in an identifiable {\em LEX} area. It should also be possible to observe the formation, on-the-fly, of assemblies corresponding to syntactic structures and parts of speech formed during parsing; one could also identify the corresponding areas (such as {\sc Verb}, {\sc Subj}, etc.). When it comes to identifying areas, the precise anatomical location of these may vary per individual, but be consistent in any individual.  Our ultimate prediction is that the long-sought {\em neural basis of language} consists of --- or rather, can be usefully abstracted as --- a collection of brain areas and neural fibers at the left MTL and STG and elsewhere in the brain, powered by the {\em project*} operation, and adapted during the critical period to an individual's maternal tongue and other circumstances.

\paragraph{Future Work.} Future work will focus on extending the scope of the Parser. This includes the extensions mentioned above (embedding and recursion in particular). 

 Another focus will be integrating this work with the existing directions in computational psycholinguistics. This includes enhancing the parser to exhibit the hallmark psycholinguistic desiderata described in Section \ref{related}. Our parser in fact has {\em incrementality} in the same sense as this literature, but it would be interesting to achieve {\em connectedness} of intermediate structures. Another future direction would be to consider how a parser implemented in AC could be used to predict or model experimental data (such as processing time). 

To conclude, we highlight one open problem for future work, the contextual action problem. We are given sentence $s$ with labeled dependencies $\mathcal{D}$, and a Parser with an area for each label that occurs in $\mathcal{D}$, as well as {\sc Lex}. Do there exist {\em contextual actions} $\alpha_w^*$ for each word $w \in s$ such that the parsing algorithm, combined with readout, yields $\mathcal{D}$? Can we construct an oracle $\mathcal{O}(s,w)$ that returns the contextual actions? If not, then what set of labeled dependency trees can be recognized under this formalism? 

How are such actions represented neurally? Can we plausibly implement contextual actions in AC, based on a word and its immediate neighbors in $s$? A step towards this may be to first implement inhibition and disinhibition of areas and fibers, treated as primitive operations in this paper, on the level of neurons and synapses (by modeling the inhibitory populations with {\em negative} edge weights, and connecting them to assemblies in {\sc Lex}).

\section{Conclusion}
Few students of the brain think seriously about language, because (1) language is by far the most advanced achievement of any brain, and (2) despite torrential progress in experimental neuroscience, an overarching understanding of how the brain works is not emerging, and the study of language will require that.  This is most unfortunate, because language has evolved rapidly over a few thousand generations, presumably to adapt to the capabilities of the human brain, and it therefore presents a great opportunity for neuroscience.  This paper is a small first step towards establishing a framework for studying, computationally, linguistic phenomena from the neuroscience perspective, and we hope that it will be followed by bolder experiments and far-reaching advancements in our understanding of how our brain makes language and the mind.

\section{Details of the experiment} \label{appendix}
\paragraph{English} We generated 10 examples each from the 20 templates shown below, a total of 200 sentences. Our Parser simulator correctly parsed all 200 examples, in the sense that a correct dependency graph was enerated by the readout.  Each of the 20 templates is a part-of-speech sequence. For each template we show below an example sentence, and in the code files we provide the correct dependencies for each to compare to the Parser's output. In the templates below, {\sc V} = {\sc V-trans}.
\begin{enumerate}[noitemsep] 
\item {\sc N V-intrans} (people died)
\item {\sc N V N} (dogs chase cats)
\item {\sc D N V-intrans} (the boy cried)
\item {\sc D N V N} or {\sc N V D N} (the kids love toys)
\item {\sc D N V D N} (the man saw the woman)
\item {\sc ADJ N V N} or {\sc N V ADJ N} (cats hate loud noises)
\item {\sc D Adj N D Adj N} (the rich man bought a fancy car)
\item {\sc Pro V Pro} (I love you)
\item {\sc \{D\} N V-intrans Adverb} (fish swim quickly)
\item {\sc \{D\} N Adverb V-intrans} (the cat gently meowed)
\item {\sc \{D\} N V-intrans Adverb} (green ideas sleep furiously)
\item {\sc  \{D\} N Adverb V \{D\} N} (the cat voraciously ate the food)
\item {\sc   \{D\} N V-intrans PP} (the boy went to school)
\item {\sc   \{D\} N V-intrans PP PP} (he went to school with the backpack)
\item {\sc  \{D\} N V \{D\} N PP  } (cats love the taste of tuna)
\item {\sc  \{D\} N PP V N} (the couple in the house saw the thief)
\item {\sc  \{D\} N copula  \{D\} N} (geese are birds)
\item {\sc  \{D\} N copula Adj} (geese are loud)
\item {\em complex sentences with copula} (big houses are expensive)
\item {\em chained adjectives, extended model} (the big bad problem is scary)
\end{enumerate}

\paragraph{Russian} To demonstrate parsing of a syntactically very different language, we consider Russian sentences with subject, object, and indirect-object, like ``\foreignlanguage{russian}{женщина дала мужчине сумку}" ({\em woman}-nom {\em give}-past {\em man}-dat {\em bag}-acc, ``the woman gave the man a bag"). All $4! = 24$ permutations of this sentence are valid, e.g. ``\foreignlanguage{russian}{сумку женщина мужчине дала}". For each of them, the Parser produces the same dependencies $\mathcal{D}=$ \{ \foreignlanguage{russian}{дала}$\xrightarrow{\textsc{Nom}}$\foreignlanguage{russian}{женщина}, \foreignlanguage{russian}{дала}$\xrightarrow{\textsc{Dat}}$\foreignlanguage{russian}{мужчине}, \foreignlanguage{russian}{дала}$\xrightarrow{\textsc{Acc}}$\foreignlanguage{russian}{сумку} \}.

\section*{Acknowledgements}
We would like to thank Dan Gildea and the anonymous reviewers for their very useful feedback. CHP and DM's research was partially supported by NSF awards CCF1763970 and CCF1910700 and by a research contract with Softbank; SSV's by NSF awards 1717349, 1839323 and 1909756; WM's by the Human Brain Project grant 785907 of the European Union; and a grant from CAIT (Columbia Center for AI Research). 

\bibliography{tacl2018}
\bibliographystyle{acl_natbib}

\end{document}